# Registration-Free Monitoring of Unstructured Point Cloud Data via Intrinsic Geometrical Properties

Mariafrancesca Patalano[1], Giovanna Capizzi[1], Kamran Paynabar[2]

[1]Department of Statistical Sciences, University of Padua
[2]School of Industrial and Systems Engineering, Georgia Institute of Technology

**Abstract**

Modern sensing technologies have enabled the collection of unstructured point cloud data (PCD) of varying sizes, which are used to monitor the geometric accuracy of 3D objects. PCD are widely applied in advanced manufacturing processes, including additive, subtractive, and hybrid manufacturing. To ensure the consistency of analysis and avoid false alarms, preprocessing steps such as registration and mesh reconstruction are commonly applied prior to monitoring. However, these steps are error-prone, time-consuming and may introduce artifacts, potentially affecting monitoring outcomes. In this paper, we present a novel registration-free approach for monitoring PCD of complex shapes, eliminating the need for both registration and mesh reconstruction. Our proposal consists of two alternative feature learning methods and a common monitoring scheme designed to handle hundreds of features. Feature learning methods leverage intrinsic geometric properties of the shape, captured via the Laplacian and geodesic distances. In the monitoring scheme, thresholding techniques are used to further select intrinsic features most indicative of potential out-of-control conditions. Numerical experiments and case studies highlight the effectiveness of the proposed approach in identifying different types of defects.

*Keywords:* Statistical Process Monitoring; Additive Manufacturing; Complex Shapes; Laplace-Beltrami Operator; Geodesic Distances.

# 1 Introduction

Recent advances in additive manufacturing (AM) have enabled the production of printed parts featuring high shape complexity, lightweight design, and topologically optimized structures. Such innovative shapes have gained widespread use across various fields, ranging from the aerospace and automotive industries to generative and prosthetic design. Monitoring the shape of printed parts to detect geometric deviations from the nominal design is essential and can be performed either retrospectively for a batch, in the absence of a reference sample of in-control (IC) parts (Phase I), or online, when a reference sample is available (Phase II). In Phase II, monitoring can be conducted ex-situ (i.e., after each part is produced) or in-situ (i.e., during the production process). To inspect printed components and assess their geometric accuracy, three-dimensional (3D) point cloud data (PCD) are collected. In some cases, structured PCD are acquired using contact sensors, such as coordinate measuring machines, which tend to be slow. The term structured refers to data organized in a grid or a regular structure. In contrast, inspection of complex shapes is typically performed using non-contact metrology systems, such as laser scanners and X-ray computed tomography (Kalender, 2011). Scanned components are subsequently monitored as high-dimensional (HD), unstructured PCD of varying sizes, even for identical parts. Each printed part and its corresponding PCD are often represented as a triangle mesh consisting of a set of vertices, edges, and faces. In contrast to raw point clouds, the mesh stores additional information, namely the connectivity among vertices. When analyzing point cloud or mesh data, the manifold assumption is typically adopted. It assumes that points lie on a low-dimensional curved space, called a manifold, that locally resembles a flat Euclidean space and is embedded in a HD ambient space. In this context, a new notion of distance is required, since the distance along the surface (i.e., geodesic distance) must be computed between two points that are far apart, rather than the Euclidean distance (see Figure 1). Developing efficient monitoring methods for unstructured PCD presents several challenges: (*i*) high-dimensionality: PCD contain thousands to millions of points; (*ii*) varying size: the number of collected points varies even for identical parts; (*iii*) lack of correspondence: unstructured PCD have no natural ordering, preventing direct point-to-point comparison; (*iv*) curved geometry: distances and features must be computed along the manifold surface rather than in the Euclidean space. To address these challenges, two common preprocessing steps are often applied: registration (i.e., preliminary alignment of the nominal and actual point clouds) and mesh reconstruction. The standard algorithm for achieving point cloud registration is Iterative Closest Point (ICP) (Besl and McKay, 1992). Nevertheless, ICP has several drawbacks, including a small convergence basin and sensitiv-ity to outliers and partial overlaps. Mesh reconstruction is another preprocessing step that requires careful attention, particularly when dealing with complex geometries. A broad range of algorithms for mesh reconstruction, commonly known as surface reconstruction, is available (see, e.g., Edelsbrunner et al., 1983; Bernardini et al., 1999; Amenta and Bern, 1999; Kazhdan et al., 2006; Kazhdan and Hoppe, 2013 for representative methods). However, reconstruction algorithms can introduce holes, artifacts, and spurious connections, leading to the loss of sharp features. In more severe cases, reconstructed meshes may even encode incorrect geometry, thereby affecting monitoring outcomes (see Supplementary Material Section S2 for further discussion). At present, monitoring varying-size unstructured point clouds without relying on registration or mesh reconstruction remains a challenging task. This highlights the need for an efficient online monitoring framework, capable of detecting both global and local defects, while avoiding critical preprocessing steps.

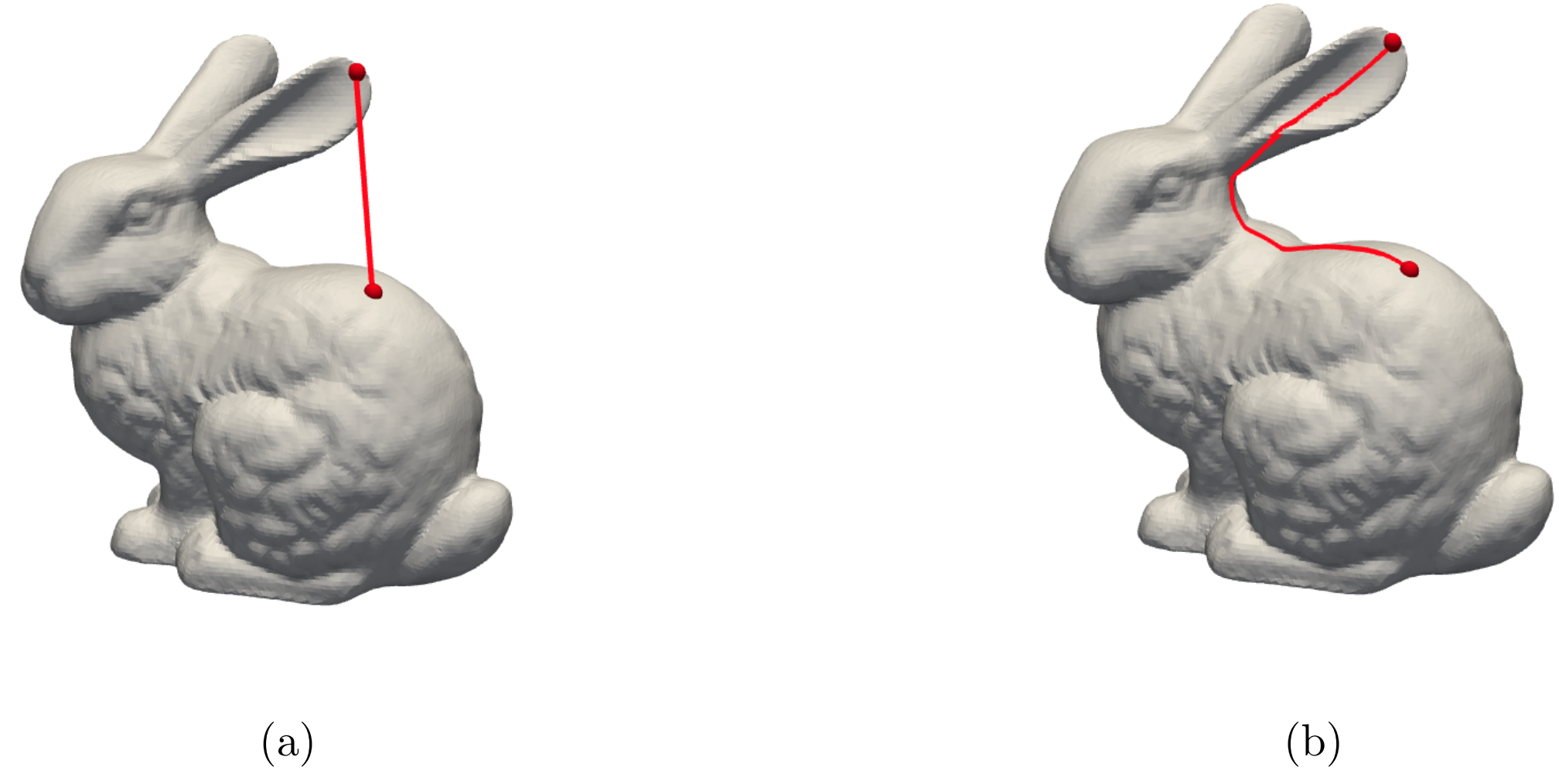

(a) (b)

Figure 1: Paths between two points: Euclidean (a) and geodesic (b).

In this paper, we propose two alternative procedures for ex-situ online monitoring of 3D printed parts. Our aim is to address the aforementioned challenges and provide a flexible solution for monitoring complex shapes from HD unstructured PCD. The main contributions are: (*i*) our procedures constitute the first monitoring methods that operate directly on raw and unaligned point clouds of varying size, eliminating both registration and mesh reconstruction; (*ii*) hundreds of geometric features are monitored simultaneously to capture fine-scale shape details and detect subtle defects; (*iii*) we introduce a novel methodology to handle the inherent high variability of geodesic distances, enabling robust monitoring. In particular, in both procedures, we use spectral dimensionality reduction techniques to capture intrinsic geometric features of the shape, while remaining independent of point cloud size. The two registration-free monitoring (RFM) procedures differ in the method used to learn the features. In the first procedure, referred to as RFM-RL, features are derived from the spectrum of the Robust Laplacian (RL) (Sharp and Crane, 2020). The second, referred to as RFM-HM, is based on geodesic distances, computed via the Heat Method (HM) (Crane et al., 2017). From the geodesic distances, a similarity matrix is derived and part of its spectrum represents the learned features. The two approaches fundamentally differ in their nature: the first captures geometrical features by aggregating information computed from local structures, whereas the second learns the underlying geometry by globally analyzing all pairwise distances. The extracted features are then employed in a monitoring scheme shared by both RFM-RL and RFM-HM. These features provide a hierarchical geometric signature: lower-order features capture coarse global shape descriptors, while higher-order features progressively encode finer local details. Thus, monitoring hundreds of features enables detection of both large-scale deformations and subtle localized defects. In the monitoring scheme, thresholding techniques are used to select intrinsic features most indicative of potential out-of-control (OC) conditions. The two proposed procedures are applicable to high-volume AM applications, where IC samples are naturally available, or low-volume production, where IC samples can be generated from the computer-aided design (CAD) model or a few IC parts using methods that estimate the IC noise distribution (see, e.g. Senin et al., 2021).

The remainder of the paper is organized as follows: Section 2 reviews related works and Section 3 presents RFM-RL and RFM-HM, introducing the feature learning methods and the monitoring scheme. Performance evaluations through simulation studies are provided in Section 4, while Section 5 includes an application to a pragmatic example involving manufactured parts. Finally, Section 6 concludes the paper.

## 2 Related Work

Before addressing the inspection of 3D shapes, approaches proposed in the literature have traditionally focused on surface data, either through profile-based methods, in which the quality characteristic to be monitored is modeled as a function of the other two coordinates (see, e.g., Wang et al., 2014; Zang and Qiu, 2018a,b), or by parameterizing Cartesian coordinates as functions of surface coordinates (Del Castillo et al., 2015). However, when monitoring full 3D shapes, surface-based methodologies become computationally expensive or even intractable due to the high-density of point cloud data (Colosimo et al., 2022).

With the aim of inspecting 3D shapes, several deviation-based methods have been suggested. Throughout this line of research, the distribution of point-to-point distances is used to compare the point cloud with the original 3D model (i.e., CAD nominal geometry). These methods include monitoring quantile-quantile plots (Wells et al., 2013), dividing point clouds into regions of interest (ROIs) Huang et al., 2018; Stankus and Castillo-Villar, 2019, and leveraging layer-wise voxelization for lattice structures (Colosimo et al., 2022, 2024). However, defining ROIs or regular voxels can be challenging or even impractical for complex shapes, due to the unstructured aspect of the point clouds. Within the class of deviation-based approaches, the method proposed by Scimone et al. (2022) is capable of inspecting extremely complex shapes, such as hollow printed objects. Their method leverages the bi-directional nature of the Hausdorff distance to compute two deviation maps. Then, the corresponding probability density functions (PDF) are monitored via simplicial functional principal component analysis (Hron et al., 2016; Menafoglio et al., 2018). The use of both maps allows different defects to be identified; specifically, one map is more sensitive to lack of material, while the other is sensitive to excess of material. Nevertheless, spatial information is lost when summarizing deviations through PDF curves. Another drawback of deviation-based methods is that they typically rely on registration.

Some approaches have been suggested to completely avoid registration by leveraging the intrinsic properties of the manifold. Intrinsic properties are those computed using coordinates defined on the manifold itself, without employing coordinates or any information from the 3D ambient space in which the manifold is embedded (Del Castillo and Zhao, 2020). Registration-free approaches are invariant under object rotations, translations, and isometric deformations. In this context, an innovative approach has been proposed by Zhao and Del Castillo (2021) for monitoring 3D shapes in both Phase I and Phase II (ex-situ). Specifically, the lower spectrum of the discrete Laplace-Beltrami (LB) operator (Li et al., 2015) is monitored. However, their approach is applicable only to closed meshes (i.e., meshes without boundary edges). To address this limitation, Zhao and Del Castillo (2022) suggested estimating the LB spectrum in the discrete setting via Finite Element Methods (FEM), achieving a more accurate estimator, hereafter referred to as LB-FEM. The LB operator and its discrete counterparts have been widely used in geometry processing and shape analysis (see, e.g., Levy, 2006; Niethammer et al., 2007; Rustamov, 2007; Reuter et al., 2009; Patane, 2017). In particular, the LB spectrum has proven to be a powerful tool as it naturally encodes geometric properties of the shape, such as area, boundary length, and number of holes (Reuter et al., 2006). To the best of our knowledge, LB-FEM represents the current state-of-the-art approach capable of inspecting open meshes without relying on registration. Nevertheless, the approach requires mesh data, which may encode incorrect geometry due to potential surface reconstruction errors. In addition, mesh data may include low-quality elements (e.g., narrow, elongated triangles), leading to approximation errors and poor numerical conditioning in FEM (see Supplementary Material Section S1. Consequently, no existing method is capable of monitoring raw point clouds while simultaneously avoiding both registration and mesh reconstruction.

# 3 Methodology

This section details the proposed registration-free approaches RFM-RL and RFM-HM for monitoring the shape accuracy using PCD. Throughout the paper, boldface capital letters (e.g., $\mathbf{X}$) denote matrices, boldface lowercase (e.g., $\boldsymbol{x}$) denote vectors, and scalars are denoted by lowercase letters (e.g., $x$). Let $\mathbf{S} \in \mathbb{R}^{n \times d}$ denote a point cloud of $n$ points with $d = 3$. We assume that a set of $m$ IC point clouds $\{\mathbf{S}_p\}_{p=1}^{m}$ is available prior to monitoring, where $\mathbf{S}_p \in \mathbb{R}^{n_p \times d}$. During online monitoring, we will denote the point cloud collected at time $t$ by $\mathbf{S}_t \in \mathbb{R}^{n_t \times d}$.

To estimate control limits, we propose a permutation-based approach that repeatedly partitions the $m$ samples into three subsets: *training* set (of size $m_1$) used to estimate the parameters that characterize the IC distribution of the features, *tuning* set (size $m_2$) to compute the control limits, and a *test* set (size $m_3$) to evaluate the attained IC performance. This permutation-based approach provides estimation uncertainty in the IC parameters, control limits, and attained IC performance. Details of the control limit estimation are provided in Section 3.3 and Appendix A.2.

In the following, the two feature learning methods are described. Subsequently, the monitoring scheme is presented, followed by details on the design of the control chart.

## 3.1 Low-dimensional Feature Learning

### 3.1.1 Robust Laplacian

The LB operator (or Laplacian) is the generalization to curved domains of the ordinary Laplacian, defined in the Euclidean space. It plays a key role in geometry processing, physics and, more broadly, in applications involving learning on non-Euclidean domains (Bronstein et al., 2017). Given a manifold $\mathcal{M}$ and a function $f : \mathcal{M} \rightarrow \mathbb{R}$, the Laplacian $\Delta$ is defined as

$$\Delta f = -\nabla \cdot \nabla f \tag{1}$$

where $\nabla$ denotes the gradient and $\nabla\cdot$ the divergence on the manifold. Generally speaking, the Laplacian measures the deviation of the function $f$ from its average in an infinitesimal neighborhood around each point. The standard mesh Laplacian, known as cotan Laplacian, is widely adopted and is mathematically equivalent to the Laplacian obtained via linear FEM. In addition, the manifold assumption is not strictly required to construct the cotan Laplacian, and non-manifold components frequently occur after surface reconstruction. Specifically, an edge is non-manifold when it is shared by more than two faces. However, the cotan Laplacian may be numerically unstable on irregular meshes, with extremely uneven triangle areas, as well as on meshes exhibiting non-manifold edges.

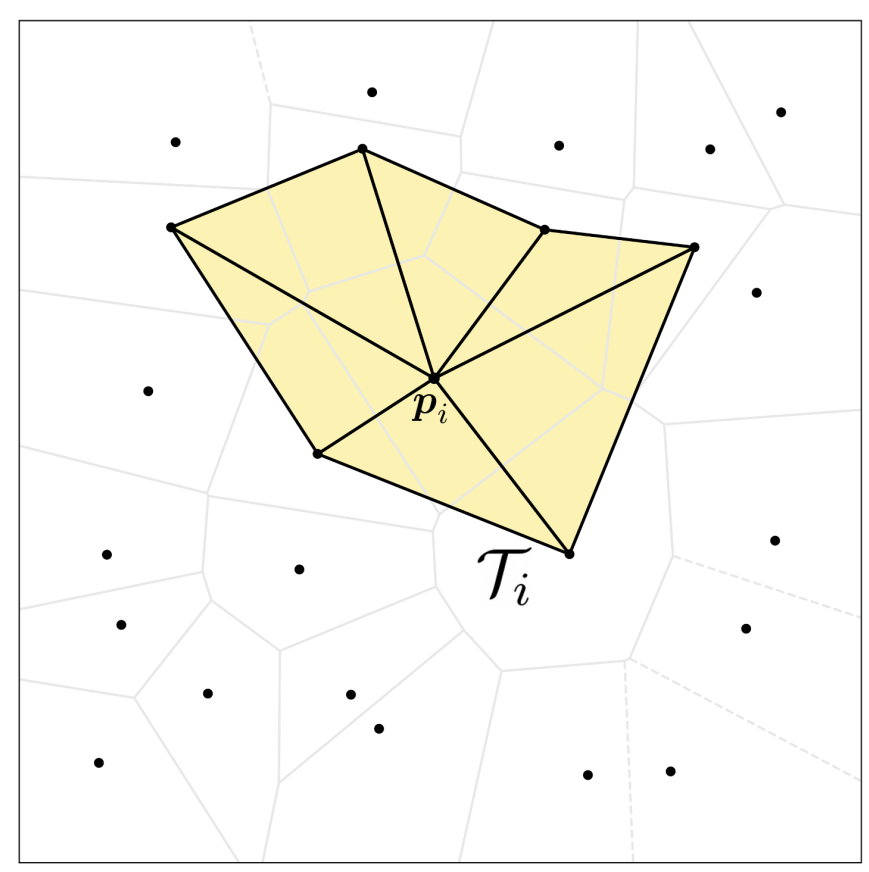

Figure 2: Local triangulation $\mathcal{T}_i$ for an interior point $\boldsymbol{p}_i$.

Sharp and Crane (2020) proposed RL, a novel discrete Laplacian suitable for PCD, even those featuring irregular sampling or non-uniform point density. Alternative point cloud Laplacians have been proposed by Clarenz et al. (2004), Belkin et al. (2009) and Liu et al. (2012). In contrast to the cotan Laplacian, RL guarantees non-negative edge weights, a desirable property for discrete Laplacians (Wardetzky et al., 2007). Analogous to the graph Laplacian, for a real-valued function defined on the vertices, non-negative edge weights ensure that the function value at each interior vertex can be expressed as a convex combination of the values at its neighbors. Hence, non-negative edge weights are desirable for both numerical stability and interpretability (see further discussion in Section S1 of the Supplementary Material).

Next, we describe the construction of RL. Let $\boldsymbol{p}_i \in \mathbb{R}^3$ denote a point in the point cloud $\mathbf{S} \in \mathbb{R}^{n\times 3}$, and $\mathcal{T}_i$ the local triangulation associated with $\boldsymbol{p}_i$ (Figure 2). To obtain RL, the triangulation $\mathcal{T} = \cup_{\boldsymbol{p}_i \in \mathbf{S}} \mathcal{T}_i$ is computed, which features highly irregular connectivity and non-manifold edges. Subsequently, for each triangle, two copies are created and glued along the edges (see Figure 3). In the resulting triangulation, all edges are manifold and no boundary edges are present, thereby enabling a well-defined discrete Laplacian. Then, the edges are flipped to obtain an intrinsic Delaunay triangulation (Bobenko and Springborn, 2007), ensuring non-negative edge weights while preserving the original geometry (Figure 4). The cotan Laplacian is then computed on the resulting triangulation, and the following generalized eigenvalue problem is solved

$$\mathbf{L}\boldsymbol{u} = \lambda \mathbf{M}\boldsymbol{u} \tag{2}$$

where $\mathbf{L}$ and $\mathbf{M}$ denote the stiffness matrix and the mass matrix, respectively. Specifically, $\mathbf{L} \in \mathbb{R}^{n\times n}$ is built by summing up entries of local matrices $\mathbf{L}_{ijk}$, defined for each triangle $(i,\, j,\, k)$ with vertices $\boldsymbol{p}_i$, $\boldsymbol{p}_j$ and $\boldsymbol{p}_k$

$$\mathbf{L}_{ijk} = \begin{bmatrix} w_j^{ki} + w_k^{ij} & -w_k^{ij} & -w_j^{ki} \\ -w_k^{ij} & w_k^{ij} + w_i^{jk} & -w_i^{jk} \\ -w_j^{ki} & -w_i^{jk} & w_j^{ki} + w_i^{jk} \end{bmatrix}$$

where $w_i^{jk} = \frac{1}{2}\cot\theta_i^{jk}$ and $\theta_i^{jk}$ is the angle at vertex $\boldsymbol{p}_i$, opposed to edge $\ell_{jk}$. The mass matrix $\mathbf{M} \in \mathbb{R}^{n\times n}$ is diagonal with entries $m_{ii} = \sum_{(i,j,k)\in\mathcal{T}_i} \mathcal{A}_{ijk}/3$, where $\mathcal{A}_{ijk}$ denotes the area of the triangle $(i,\, j,\, k)$. That is, each triangle contributes one-third of its area to the mass associated with each of its vertices (see Sharp and Crane (2020) for further details).

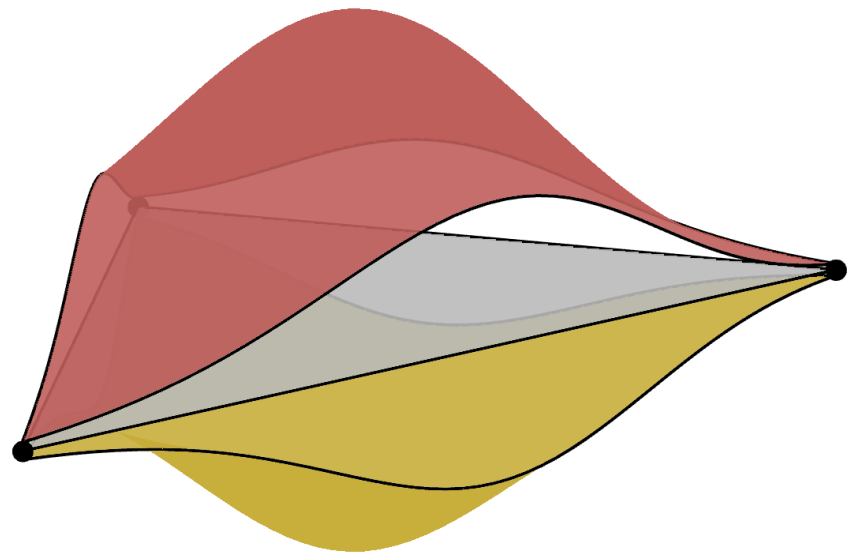

Figure 3: Two copies of each triangle.

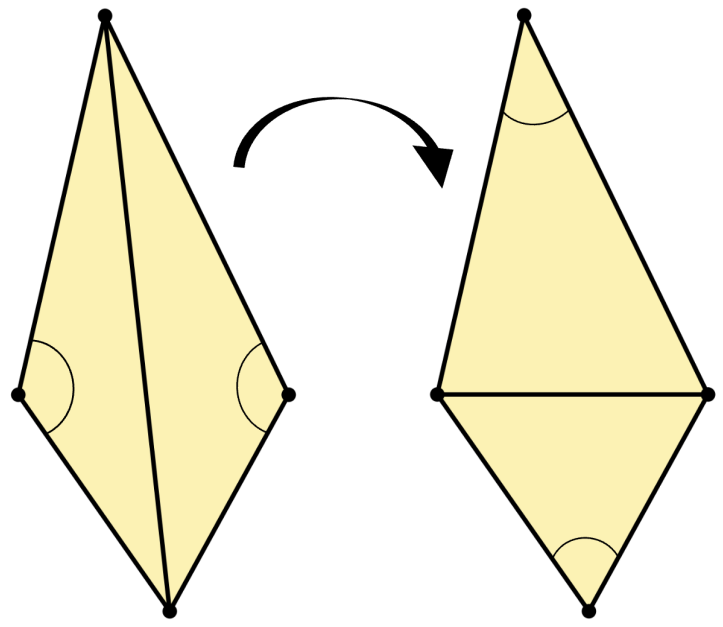

Figure 4: Edge flip.

In our RFM-RL procedure, the eigenvalue problem in Eq. (2) is solved to obtain the spectrum corresponding to the point cloud $\mathbf{S} \in \mathbb{R}^{n\times 3}$. By definition, the first eigenvalue is zero and is customarily omitted from subsequent analyses. In our setting, as in most applications, the lower spectrum $\boldsymbol{\lambda} \in \mathbb{R}^k$ $(k << n)$ is considered, as it captures global shape descriptors, while higher eigenvalues are generally associated with fine-scale features and are more sensitive to noise. Consequently, we solve a truncated eigenvalue problem, computed efficiently through the ARPACK library (Lehoucq et al., 1998).

#### 3.1.2 Geodesic Distances via Heat Method

The geodesic distance between two points on a surface is the length of the shortest path connecting them while being constrained to the surface (Figure 1). Our proposal leverages all pairwise geodesic distances to capture relationships between points across multiple scales. In the literature, several methods have been developed to compute geodesic distances, including graph-based methods (Dijkstra, 1959; Lanthier et al., 1997), exact polyhedral schemes (Mitchell et al., 1987; Surazhsky et al., 2005), and methods based on partial differential equations (PDE), such as Fast Marching (Kimmel and Sethian, 1998), and spectral distances (see Patane, 2017 for a review). Most of them are developed for triangle meshes, only few contributions have been designed for point clouds, (see, e.g., Tenenbaum et al., 2000; Bernstein et al., 2000; Mémoli and Sapiro, 2005). Refer to Crane et al., 2020 for a comprehensive review of algorithms for geodesic distances.

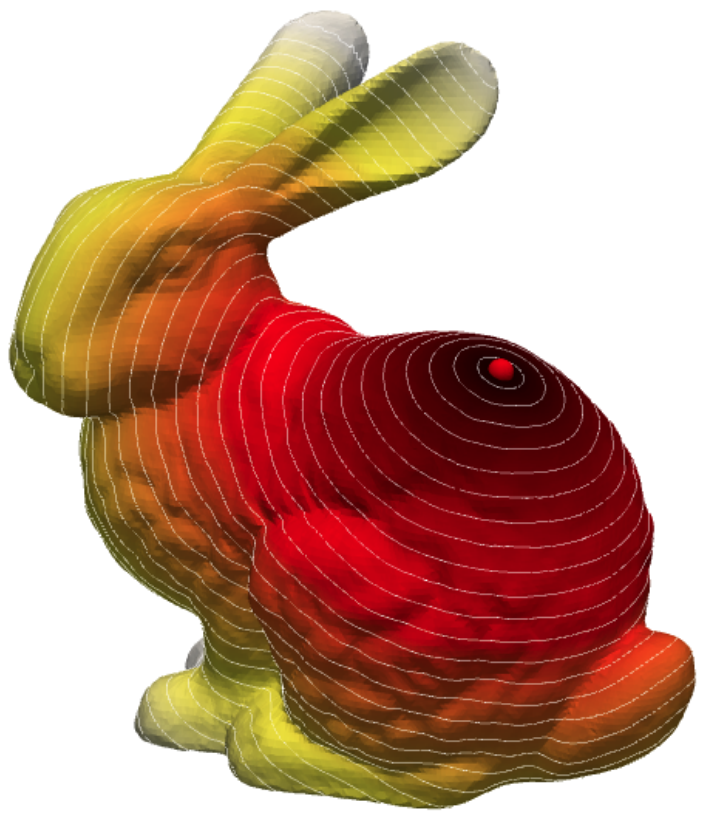

Figure 5: Geodesic distance from a single source point on a surface.

Inspired by the relationship between heat diffusion and geodesic distances, HM (Crane et al., 2017) can be used to efficiently compute geodesic distances. HM belongs to the class of PDE-based methods and is applicable to a wide variety of data structures, including point clouds. The heat transferred from a source point to another location over a short time is closely related to the geodesic distance between the two points. Hence, HM leverages short-time heat diffusion to compute geodesic distances. Figure 5 depicts an object and the heat spread across its surface over time, starting from a single source point.

Formally, let $\phi : \mathbb{R}^d \times \mathbb{R}^d \to \mathbb{R}$ be the geodesic distance between a pair of points and $v_t : \mathbb{R}^d \to \mathbb{R}$ a radially symmetric function that decreases monotonically with distance. In practice, $v_t$ is an approximation of the heat flow for a fixed time $t$. The key idea of HM is to recover $\phi$ by finding the function $v_t$ whose gradient $\nabla v_t$ is parallel to $\nabla\phi$. Specifically, three primary steps are involved: $(i)$ solve the heat equation $\frac{d}{dt}v + \Delta v = 0$ for $t$ fixed, with a Dirac delta centered at the source point as the initial condition, $(ii)$ evaluate the vector field $X = -\nabla v_t / |\nabla v_t|$ and $(iii)$ solve the Poisson equation $\Delta\phi + \nabla \cdot X = 0$. Note that the heat and Poisson equations are stated according to the definition of the Laplacian in Eq. (1). Time and space discretizations are adopted to apply this three-step procedure to the discrete setting (see Crane et al., 2017 for further details).

Using HM, we obtain the matrix of all pairwise geodesic distances $\mathbf{D} \in \mathbb{R}^{n\times n}$ for the point cloud $\mathbf{S} \in \mathbb{R}^{n\times 3}$. In RFM-HM, we derive the similarity matrix $\mathbf{B} \in \mathbb{R}^{n\times n}$ as follows

$$\mathbf{B} = -\frac{1}{2}\,\mathbf{H}\,\mathbf{D}\,\mathbf{H} \tag{3}$$

where $\mathbf{H}$ is the centering matrix, $\mathbf{H} = \mathbf{I} - \frac{1}{n}\,\mathbf{1}\mathbf{1}^\top$. In Eq. (3), we propose a modification of classical multidimensional scaling (CMDS), by using $\mathbf{D}$ instead of $\mathbf{D}^2$, thus applying a variance-stabilizing transformation. The eigendecomposition of $\mathbf{B}$ is computed and the leading part of the spectrum represents the extracted feature vector $\boldsymbol{\lambda} \in \mathbb{R}^k$ for the point cloud $\mathbf{S} \in \mathbb{R}^{n\times 3}$. The truncated eigenvalue problem is solved using ARPACK, whereas we recommend solving the full eigenvalue problem on a small subset of the reference sample to select the appropriate number of eigenvalues (see Section 3.3).

## 3.2 Monitoring Scheme

By applying one of the two proposed feature learning methods, a feature vector $\boldsymbol{\lambda} \in \mathbb{R}^k$ is obtained from the point cloud $\mathbf{S} \in \mathbb{R}^{n\times 3}$. From the reference sample, feature vectors are computed equivalently, resulting in a set of $m$ IC realizations. We assume that the IC feature vectors are drawn from a generic distribution $F(\cdot)$, with mean $\boldsymbol{\mu} \in \mathbb{R}^k$ and covariance matrix $\boldsymbol{\Sigma} \in \mathbb{R}^{k\times k}$.

When monitoring begins, at time $t$, the point cloud $\mathbf{S}_t \in \mathbb{R}^{n_t\times 3}$ is collected and the corresponding feature vector $\boldsymbol{\lambda}_t \in \mathbb{R}^k$ is obtained. Our objective is to rapidly detect any undesirable condition, shape deformation, or deviation from the nominal design. To cover a wide range of defect types, we formulate the problem as detecting changes in the mean of the feature vectors over time. No specific assumptions are imposed on the magnitude, direction of the change, or number of affected components. Indeed, the OC condition may affect only a few components of the mean vector.

The selected $k-$dimensional feature vector is typically quite large. For instance, hundreds of eigenvalues may be required to capture significant features of the shape under consideration. To focus on the most informative features for detecting potential OC conditions, shrinkage methods are applied to reduce the feature set. In literature, shrinkage techniques play a relevant and well-known role in dimensionality reduction and feature selection and have been successfully applied to profile monitoring (see, e.g., Jeong et al., 2006, 2007; Zou et al., 2010; Wang et al., 2018). Such further selection improves the power of the test, particularly when the OC condition is primarily reflected in either a sparse subset of components or in the last components (Donoho and Johnstone, 1994; Fan, 1996).

In our monitoring scheme, we introduce a control statistic based on the Mahalanobis distance (Mahalanobis, 1936) with hard-thresholding. Let $\boldsymbol{z}_t = \boldsymbol{\Sigma}^{-1/2}\left(\boldsymbol{\lambda}_t - \boldsymbol{\mu}\right) \in \mathbb{R}^k$ denote the standardized observation collected at time $t$, where $\boldsymbol{\mu}$ and $\boldsymbol{\Sigma}^{-1/2}$ are the mean and inverse square root of the covariance matrix estimated from IC observations. The control statistic is defined as

$$T_{H,c} = H(\boldsymbol{z}_t,\, c)^\top H(\boldsymbol{z}_t,\, c)$$

where $H(\boldsymbol{z}_t, c)$ is the hard-thresholding operator (Donoho and Johnstone, 1994)

$$H(\boldsymbol{z}_t, c) = \boldsymbol{z}_t \cdot \mathbb{I}\left(|\boldsymbol{z}_t| > c\right)$$

with threshold parameter $c = \sqrt{2\log(k\, a_k)}$, where $a_k = (\log k)^{-2}$. This choice for the threshold parameter is suggested by Fan (1996) and is specifically designed for feature selection in HD settings where $k$ is large.

## 3.3 Design of the Control Chart

The design of RFM-RL and RFM-HM involves the choice of the number of eigenvalues $k$, the size of the *training* and *tuning* sets, and the control limits.

The number of eigenvalues $k$, for both RFM-RL and RFM-HM, can be selected using a small number of IC parts. For RFM-RL, $k$ can be chosen according to the elbow-rule procedure described in the Appendix A.1. Since we are using thresholding techniques in the monitoring scheme, it is generally acceptable to select a slightly larger $k$ than the value indicated by the procedure. To select $k$ for RFM-HM, the cumulative explained variance criterion can be used, as in classical principal component analysis. Here, $k$ is chosen as the smallest number of eigenvalues that explain 95% of the total variance. The procedure can be applied to a subset or to all point clouds in the *training* set, and the largest value obtained can be selected.

The *training* and *tuning* sets are used to estimate $\boldsymbol{\mu}$ and $\boldsymbol{\Sigma}$ and to compute the control limits, respectively. Based on our empirical findings, we recommend $m_1 = 500$ and $m_2 = 1000$ for reliable estimation of IC parameters and control limits. The algorithm used to estimate the control limits is detailed in the Appendix A.2.

# 4 Simulation Study

## 4.1 Comparison of methods: IC and OC Performance

In this section, RFM-RL and RFM-HM are compared with benchmark methods to assess their effectiveness when global or local defects occur. In particular, four OC scenarios are considered: global and local surface roughness, scattered surface imperfections, and excess of material. The Stanford Bunny[1], a widely used mesh in computer graphics, is used for evaluation. The maximum height and width of the Bunny are approximately 0.15 mm, making it a representative example of high-resolution meshes. The mesh consists of 8146 vertices and 16301 faces, including non-manifold edges and holes, primarily at the bottom. These characteristics reflect real-world inaccuracies often encountered during scanning and reconstruction. In the following, we consider only the point cloud of the Bunny (Figure 6). To create the reference sample, noisy versions of the original point cloud are generated assuming isotropic noise $N_3(\mathbf{0}, \sigma_0^2\,\mathbf{I})$ with $\sigma_0 = 0.0001$. Simulated point clouds are generated with varying sizes, ranging between 8100 and 8140 points, by discarding a random number of points from the original point cloud. Additional results for different point densities are reported in the Supplementary Material (Section S5). Performance is assessed in terms of Average Run Length (ARL), where the run length is defined as the number of observations before an alarm is triggered. In all tables, average ARL (AARL) values are reported, with standard deviation given in parentheses. The IC ARL ($ARL_0$) is set to 20, and the maximum run length is set to 1000 and 500 to assess IC and OC performance, respectively.

[1]Data available at the Stanford 3D Scanning Repository.

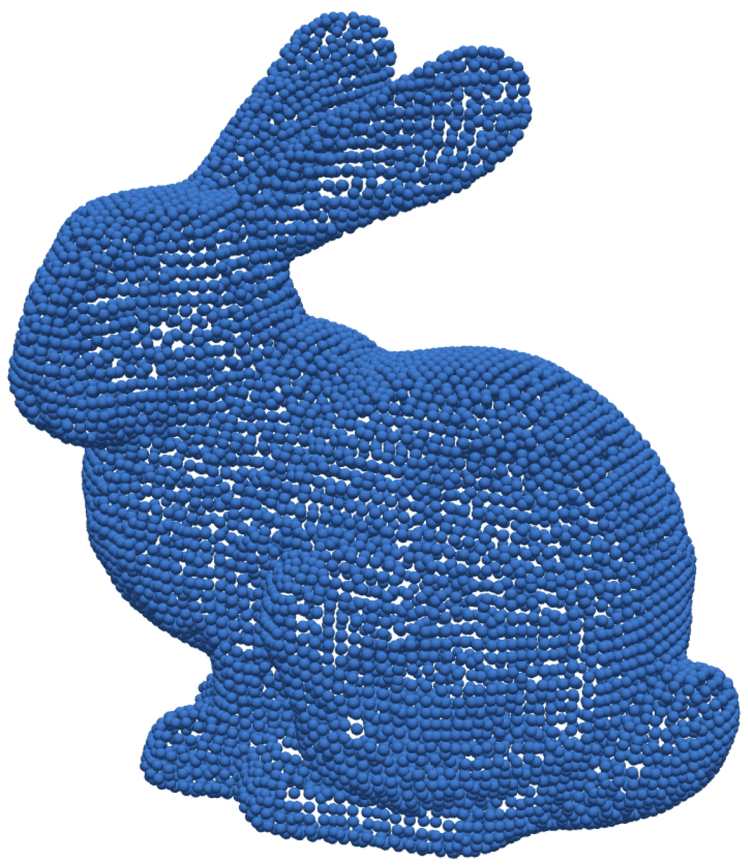

Figure 6: Point cloud of the Stanford Bunny.

The performance is evaluated including four benchmarks: ISOMAP (Tenenbaum et al., 2000), Locally Linear Embedding (LLE) (Roweis and Saul, 2000), T-distributed Stochastic Neighbor Embedding (TSNE) (van der Maaten and Hinton, 2008), and Graph Laplacian (GL) (Chung, 1997). The first three methods belong to the manifold learning literature, while the fourth is included because of its interesting analogy with the LB operator (see Chung, 1997 for further details). Manifold learning techniques aim to find low-dimensional structures hidden in the HD ambient space, and can be applied to monitor unstructured point clouds without relying on registration or mesh reconstruction. Each technique is applied to map data points from the original ambient space $\mathbb{R}^3$ to $\mathbb{R}^2$. In this embedded space, all pairwise distances are computed to approximate geodesic distances in the original space. Subsequently, CMDS is adopted to derive the similarity matrix and its truncated eigendecomposition is computed. Finally, our proposed monitoring scheme is applied. The fourth benchmark can be applied using a procedure analogous to that employed for RFM-RL. Hence, the lower spectrum is computed and monitored using our monitoring scheme. Throughout all experiments, we used 30 nearest neighbors for the construction of RL, GL, ISOMAP and LLE, following the recommendation in Sharp and Crane (2020) and keeping this value consistent for fair comparison. Notably, RL does not require kernel-scale selection or additional tuning parameters, reducing sensitivity to parameter choice.

Table 1: IC performance. Isotropic noise $N_3(\mathbf{0}, \sigma_0^2\,\mathbf{I})$, $\sigma_0 = 0.0001$. $\mathrm{ARL}_0 = 20$.

| Method | IC |
|---|---|
| ISOMAP | 20.32 (4.25) |
| LLE | 20.74 (4.24) |
| TSNE | 20.42 (4.08) |
| GL | 20.34 (3.96) |
| RFM-RL | 20.49 (4.19) |
| RFM-HM | 20.30 (4.21) |

The number of eigenvalues $k$ is chosen as 151 for RFM-HM and 100 for RFM-RL. The sizes of the *training*, *tuning*, and *test* sets are $m_1 = 500$, $m_2 = 1000$, $m_3 = 1000$, respectively.

Table 2: Global surface roughness. Isotropic IC noise $N_3(\mathbf{0}, \sigma_0^2 \mathbf{I})$, $\sigma_0 = 0.0001$ and SNR $= \sigma_1/\sigma_0$. $\text{ARL}_0 = 20$.

| | SNR | | | | | |
|---|---|---|---|---|---|---|
| Method | 1.05 | 1.15 | 1.25 | 1.35 | 1.45 | 1.55 |
| ISOMAP | 24.61 (1.98) | 21.09 (1.86) | 19.37 (1.91) | 17.66 (1.89) | 15.46 (1.33) | 13.74 (1.16) |
| LLE | 21.98 (3.04) | 25.67 (4.67) | 23.95 (3.69) | 25.72 (3.52) | 26.85 (4.68) | 32.02 (6.89) |
| TSNE | 17.59 (1.45) | 16.83 (1.66) | 14.24 (1.48) | 17.16 (1.68) | 15.49 (1.19) | 13.17 (0.76) |
| GL | 17.06 (2.33) | 13.86 (1.71) | 11.63 (1.28) | 9.99 (1.03) | 8.17 (0.81) | 6.74 (0.63) |
| RFM-RL | 16.53 (2.09) | 8.00 (1.05) | **2.73 (0.32)** | **1.26 (0.07)** | **1.01 (0.01)** | **1.00 (0.00)** |
| RFM-HM | **13.78 (2.33)** | **7.59 (0.97)** | 4.42 (0.43) | 2.68 (0.21) | 1.76 (0.09) | 1.35 (0.05) |

In Table 1, the IC performance is reported, showing that $\text{ARL}_0$ is attained for all methods, thus confirming that the Type I error is controlled at the nominal level. Table 2 presents the OC performance when global surface roughness exceeds specifications, representing uniform degradation across the entire part surface. We simulate this global defect by adding noise sampled from $N_3(\mathbf{0}, \sigma_1^2 \mathbf{I})$. We report the results in terms of signal-to-noise ratio (SNR), defined as SNR $= \sigma_1/\sigma_0$. In Table 2, RFM-HM achieves the best performance for low SNR values, while RFM-RL performs better for higher SNR. Notably, both proposed methods exhibit comparable performance across the considered range. Among the benchmarks, TSNE and GL show superior performance. However, their detection power is lower compared to RFM-HM and RFM-RL, especially for severe defects. Additional results for anisotropic surface roughness are provided in Section S4 of the Supplementary Material. In Table 3, the performance is reported when scattered surface imperfections occur at spatially random locations. This scenario represents isolated defects such as material spatter, dust contamination during printing, or random measurement outliers. We simulated this by applying noise sampled from $N_3(\mathbf{0}, \sigma_1^2\,\mathbf{I})$ with SNR $= 1.6$ to randomly selected points. The percentage of affected points, denoted by $p$, varies from 20% to 60%. Similar to the previous scenario, RFM-RL outperforms all other methods for all the values of $p$, showing high sensitivity across the range of values. RFM-HM also exhibits competitive performance, achieving the second-best results among all methods. Additionally, we consider two types of localized geometric defects. In Table 4, results are reported when localized surface roughness occurs, where surface quality degrades in a small region due to phenomena such as localized thermal gradients or cooling variations. This defect is simulated by adding isotropic noise with SNR $= 3$ to a cluster of neighboring points, and the proportion of affected points is gradually increased. In Table 5, results are reported when excess of material occurs, characterized by an additive geometric displacement resembling material over-extrusion or localized thermal expansion. This is simulated by applying a radially-symmetric shift to 97 points (approximately 1% of the total), creating a bell-shaped protrusion. The performance is evaluated for several values of the shift size (SS). In both localized defect scenarios, RFM-HM achieves the best performance, while RFM-RL achieves the second-best results among all methods.

Table 3: Scattered surface imperfections. Isotropic IC noise $N_3(\mathbf{0}, \sigma_0^2\,\mathbf{I})$, $\sigma_0 = 0.0001$, and SNR $= 1.6$. $\text{ARL}_0 = 20$.

| | $p$ | | | | |
|---|---|---|---|---|---|
| Method | 20% | 30% | 40% | 50% | 60% |
| ISOMAP | 16.46 (1.03) | 19.38 (1.85) | 14.74 (1.17) | 15.53 (1.93) | 13.65 (1.05) |
| LLE | 29.78 (5.52) | 17.09 (1.40) | 23.82 (2.55) | 21.99 (3.37) | 29.53 (4.16) |
| TSNE | 17.88 (2.00) | 16.41 (1.72) | 14.52 (1.24) | 15.41 (1.72) | 16.37 (1.20) |
| GL | 17.70 (2.55) | 14.78 (2.13) | 14.12 (1.66) | 13.23 (1.70) | 11.88 (1.47) |
| RFM-RL | **7.99 (1.01)** | **3.86 (0.47)** | **2.26 (0.22)** | **1.37 (0.08)** | **1.11 (0.03)** |
| RFM-HM | 8.49 (1.07) | 5.34 (0.53) | 4.05 (0.35) | 3.07 (0.24) | 2.35 (0.14) |

Table 4: Local surface roughness. Isotropic IC noise $N_3(\mathbf{0}, \sigma_0^2\,\mathbf{I})$, $\sigma_0 = 0.0001$ and SNR = 3. $\text{ARL}_0 = 20$.

| | $p$ | | | | |
|---|---|---|---|---|---|
| Method | 2.5% | 3.5% | 5% | 7% | 10% |
| ISOMAP | 28.64 (3.96) | 22.80 (3.05) | 23.68 (2.46) | 29.60 (2.85) | 21.62 (3.52) |
| LLE | 31.31 (5.05) | 13.45 (1.13) | 27.48 (3.05) | 41.13 (9.13) | 34.41 (5.14) |
| TSNE | 17.29 (2.21) | 22.24 (2.57) | 20.10 (1.49) | 16.59 (1.32) | 16.73 (1.47) |
| GL | 17.12 (2.50) | 16.44 (2.47) | 16.99 (2.36) | 13.35 (1.67) | 11.81 (1.42) |
| RFM-RL | 9.15 (1.71) | 4.64 (0.78) | 1.98 (0.21) | 1.18 (0.05) | 1.02 (0.01) |
| RFM-HM | **3.35 (0.39)** | **1.70 (0.13)** | **1.16 (0.04)** | **1.03 (0.01)** | **1.01 (0.00)** |

Across all evaluated scenarios, the two proposed approaches demonstrate consistently superior and reliable performance. Notably, RFM-HM exhibits the best performance in most scenarios and is closely comparable to the best method in the remaining cases. Among the benchmarks, GL generally outperforms ISOMAP, LLE and TSNE. Nevertheless, across all defect types and severity levels, none of the benchmarks show competitive performance compared to the proposed approaches. The key limitation shared by all manifold learning techniques is that pairwise distances represent only an approximation of the true geodesic distances between points. In addition, GL fails to encode geometric information, in contrast to RFM-RL, since RL, as a discrete version of the LB operator, is intrinsically geometry-aware (Rustamov, 2007).

Table 5: Excess of material. Shift of size SS applied to the center of the cluster and progressively decreased for neighboring points. Isotropic IC noise $N_3(\mathbf{0}, \sigma_0^2\,\mathbf{I})$, $\sigma_0 = 0.0001$. $\text{ARL}_0 = 20$.

| | SS | | | | |
|---|---|---|---|---|---|
| Method | 0.001 | 0.0015 | 0.002 | 0.0025 | 0.003 |
| ISOMAP | 28.08 (3.18) | 28.66 (3.44) | 29.41 (3.84) | 30.61 (4.06) | 32.08 (4.46) |
| LLE | 20.27 (2.64) | 27.90 (3.61) | 33.63 (7.32) | 26.65 (3.86) | 23.64 (2.92) |
| TSNE | 16.34 (1.10) | 13.95 (1.26) | 16.43 (1.61) | 17.45 (1.85) | 15.60 (1.94) |
| GL | 19.08 (2.93) | 18.61 (2.93) | 17.30 (2.62) | 14.89 (2.16) | 11.97 (1.74) |
| RFM-RL | 16.16 (2.21) | 8.92 (1.38) | 3.10 (0.46) | 1.28 (0.08) | **1.01 (0.01)** |
| RFM-HM | **11.89 (2.15)** | **5.32 (0.75)** | **2.11 (0.22)** | **1.16 (0.05)** | **1.01 (0.01)** |

## 4.2 Sensitivity to Small-Scale Defects and Robustness to IC noise

In this section, the performance of RFM-RL and RFM-HM is evaluated under challenging conditions involving mild global defects and severe localized defects affecting 1% of points. Local defects include local surface roughness, lack of material and excess of material (i.e., additive shift). In this section, we use the point cloud of a prosthetic aortic valve (Figure 7). 3D-printed heart valves are now increasingly being adopted for their functional perfor-mance, customization and biocompatibility (Haghiashtiani et al., 2020; Ertas et al., 2025). This application represents a crucial example of a high-resolution printing scenario in which identifying even small or mild defects is essential to ensure the reliability and functionality of the artificial valve. Moreover, mesh reconstruction is particularly non-trivial in this context (see Supplementary Material Section S2).

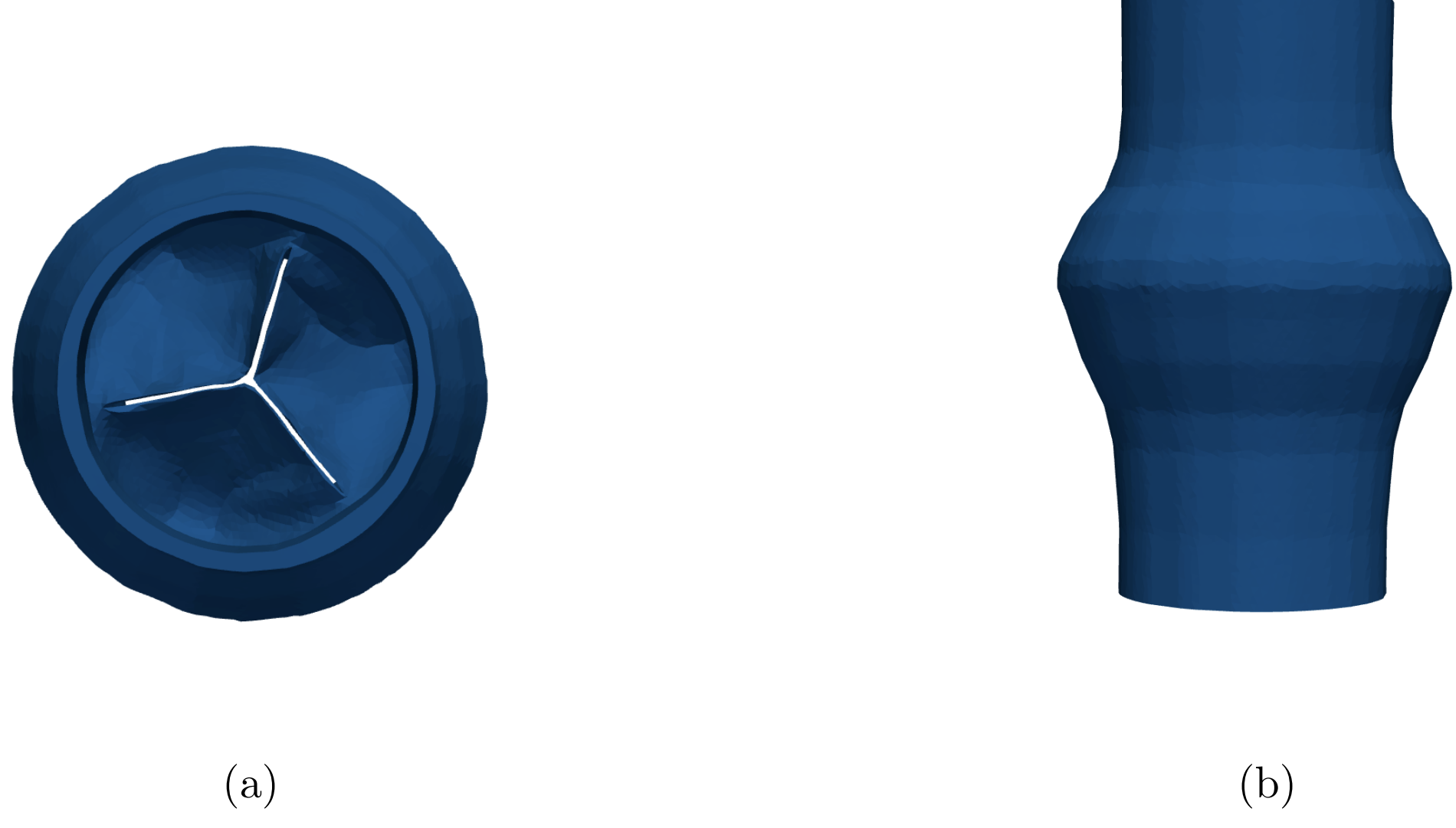

(a) (b)

Figure 7: Aortic valve: frontal view (a) and side view (b).

The point cloud of the original model consists of 21308 vertices[2], with a maximum height of 0.18 mm and a maximum width of approximately 0.12 mm. Simulated IC and OC noisy point clouds have varying sizes between 21250 and 21300 points, with points randomly selected and discarded. Samples of the *training*, *tuning* and *test* sets are generated using isotropic noise $N_3(\mathbf{0}, \sigma_0^2\,\mathbf{I})$. To assess robustness, two IC noise levels are considered ($\sigma_0 = 0.0001$ and $\sigma_0 = 0.0003$). The sample sizes are set to $m_1 = 500$, $m_2 = 1000$, and $m_3 = 1000$ for both RFM-RL and RFM-HM and $\text{ARL}_0 = 20$. The maximum run length is set to 1000 and 500 to assess IC and OC performance, respectively. The number of eigenvalues $k$ is selected using the proposed elbow rule for RFM-RL, yielding $k = 90$ for $\sigma_0 = 0.0001$ and $k = 120$ for $\sigma_0 = 0.0003$. For RFM-HM, $k = 343$ and $k = 341$ for $\sigma_0 = 0.0001$ and $\sigma_0 = 0.0003$, respectively. This indicates that increasing the noise level has little or no impact on the cumulative variance explained with the first $k$ eigenvalues. In this section, we omit the results of the four benchmarks, as they already demonstrated relatively poor performance on moderate-to-severe defects and are not competitive for the small-scale defects considered here. In the presence of severe local defects, RFM-RL exhibits higher sensitivity and consistently outperforms RFM-HM. Furthermore, RFM-RL proves to be robust, while the detection performance of RFM-HM deteriorates as $\sigma_0$ increases. However, notably, both methods demonstrate effective performance in the lack of material scenario, where a region of the part is absent. Such detection power arises primarily because the lower spectrum of LB is related to the area of the object, which decreases when material is missing, while for RFM-HM, although lack occurs locally, all points are affected since all distances are reduced, meaning that geodesic paths between points are systematically shorter than expected.

[2]Data available here. The original model includes 2377 vertices and 4754 faces, and its point cloud is structured and sparsely sampled. To obtain an unstructured point cloud, remeshing is performed using the filter available in the Python library `pymeshlab` (Hoppe et al., 1993; Muntoni et al., 2025)

Table 6: Mild global and severe local defects applied to the prosthetic valve. Isotropic IC noise $N_3(\mathbf{0}, \sigma_0^2\,\mathbf{I})$. $\text{ARL}_0 = 20$. Global surface roughness (SNR = 1.15 when $\sigma_0 = 0.0001$ and SNR = 1.05 when $\sigma_0 = 0.0003$). Local surface roughness(SNR = 5 when $\sigma_0 = 0.0001$ and SNR = 2.33 when $\sigma_0 = 0.0003$). Excess of material with SS = 0.003. Lack of material (1% of points removed).

| $\sigma_0$ | Method | IC | Local Roughness | Global Roughness | Lack | Excess |
|---|---|---|---|---|---|---|
| 0.0001 | RFM-RL | 20.24 (3.99) | **6.60 (0.86)** | **7.87 (0.91)** | **1.00 (0.00)** | **1.13 (0.03)** |
| | RFM-HM | 20.41 (4.27) | 22.32 (5.67) | 15.29 (3.11) | **1.00 (0.00)** | 4.08 (0.67) |
| 0.0003 | RFM-RL | 20.58 (4.14) | **10.81 (1.64)** | **10.05 (1.53)** | **1.00 (0.00)** | **2.59 (0.22)** |
| | RFM-HM | 20.55 (4.31) | 20.89 (5.18) | 18.66 (4.41) | **1.00 (0.00)** | 7.85 (1.39) |

As a general guideline, we suggest adopting RFM-HM for high-resolution printing applications, where the OC condition may likely affect at least 2–3% of points. In contrast, RFM-RL is better suited for detecting extremely localized and subtle defects when higher IC noise levels are specified by practitioners. Furthermore, RFM-RL is computationally more efficient (see Supplementary Material Section S3).

# 5 Case Study

This section presents the application of the two proposed approaches to a representative case study involving a component produced via AM. The case study focuses on the egg-shaped trabecular shell used in Scimone et al. (2022), hereinafter referred to as the egg shell (Figure 8a). The egg shell is a practical example of a printed part featuring lightweight design and complex free-form geometry. The original egg shells have a height of 60 mm and a maximum diameter of 40 mm, and were manufactured using polymer fused deposition modeling, where a thermoplastic filament is heated, extruded through a nozzle, and deposited layer by layer on a build platform (Mwema and Akinlabi, 2020). Here, the CAD model is uniformly scaled by a factor of 0.01, since the proposed methods require high-density point clouds.

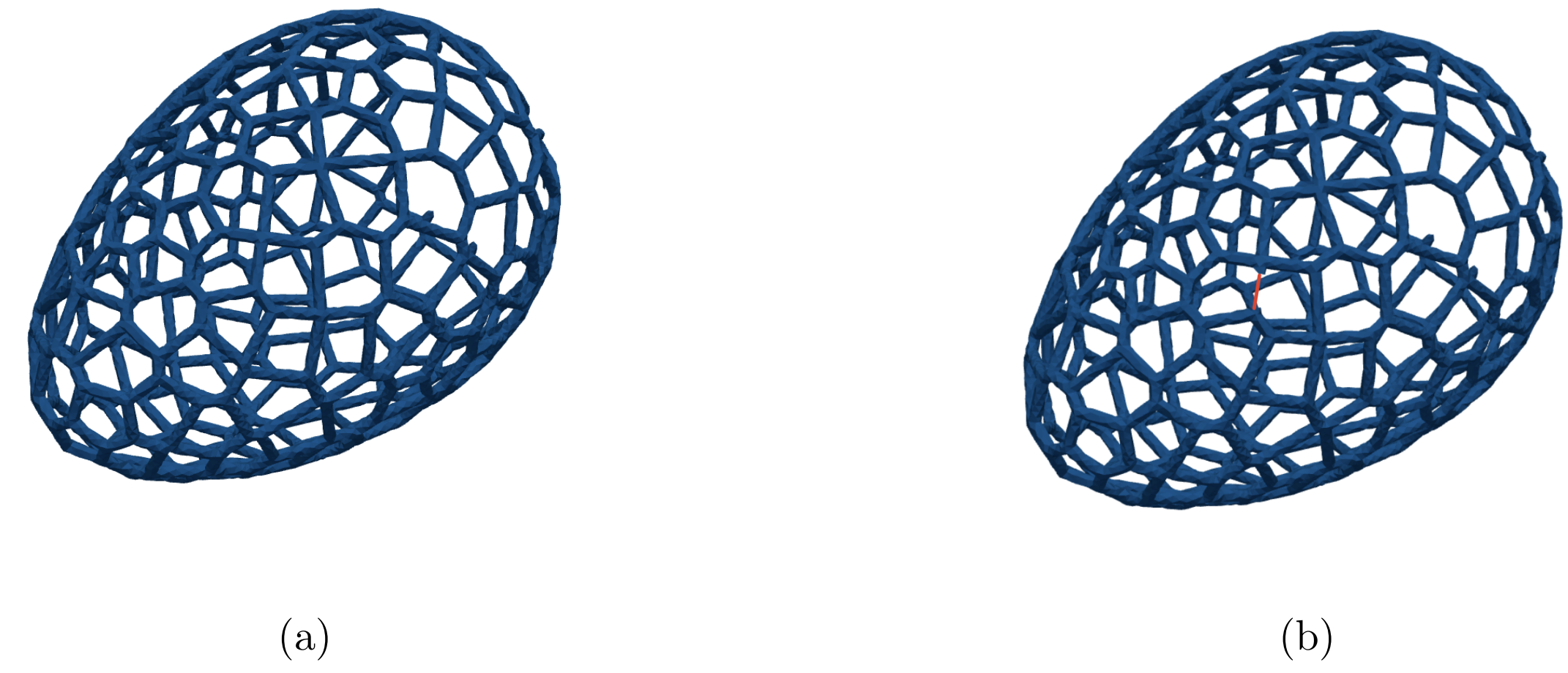

Figure 8: Egg shells: CAD model (a) and defective shell with missing strut indicated in red (b).

The aim of this section is to further validate the proposed methods and assess their effectiveness when a localized defect occurs in a printed part featuring high shape complexity. The original CAD model consists of 20003 vertices and 40718 faces. Noisy point clouds, with size varying between 19950 and 20000 points, are generated by adding isotropic noise $N_3(\mathbf{0}, \sigma_0^2\,\mathbf{I})$ with $\sigma_0 = 0.0001$. The $\mathrm{ARL}_0$ is set to 100, and $m_1$ and $m_2$ are set to 500 and 1000, respectively. For RFM-HM $k = 193$, while for RFM-RL $k = 110$. The control limit is denoted as $h$ and an alarm is raised whenever $T_{H,c} > h$. The control limit values are 127.42 for RFM-RL and 276.14 for RFM-HM. We consider the lack of material scenario involving a missing strut of the egg shell, which corresponds to 28 missing points (Figure 8b). This defect represents a topology change, as the connectivity of the egg shell is altered by the missing strut. In the sample considered, the first ten egg shells are representative of an IC condition, while the subsequent ten shells are defective due to the missing strut. Both methods are capable of detecting the OC condition, correctly identifying all OC samples

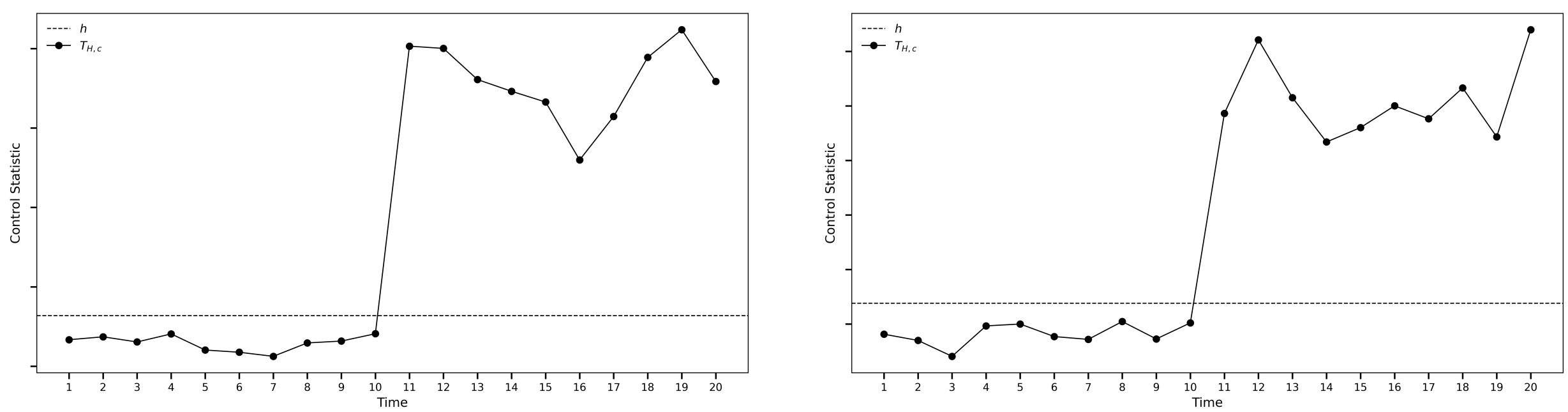

Figure 9: Control charts for egg shells monitoring using RFM-RL (left) and RFM-HM (right).

(Figure 9).

# 6 Conclusions

Assessing shape accuracy of printed parts is a challenging task as traditional monitoring techniques are not directly applicable to unstructured varying-size point clouds and existing methods rely on critical preprocessing steps. In this article, we propose a novel monitoring approach to inspect complex shapes while avoiding both registration and mesh reconstruction. Two alternative approaches are proposed to learn a low-dimensional representation of each point cloud by leveraging intrinsic properties of the shape. The most significant features are then selected to enable timely detection of a wide range of shape defects. In essence, the two approaches address feature learning from complementary perspectives: the first leverages local neighborhood information, while the second captures shape descriptors using all pairwise geodesic distances. Combining features from both approaches is not pursued, as it would require monitoring an extremely HD concatenated feature vector without offering substantial advantage, given that both methods are theoretically related through the heat kernel.

The performance of the two approaches has been investigated across various shapes and defect types. Experimental results demonstrate the ability of the proposed methods to effectively detect both global defects and localized geometric deviations in printed parts, even those featuring a high level of shape complexity. The proposed methods substantially outperform benchmark methods, particularly for subtle defects, which we attribute to the HD extracted features that capture a rich geometric signature. Results demonstrate competitive or superior performance of the method based on geodesic distances when the defect affects at least 2-3% of points, while the approach based on the Laplacian is more robust to higher levels of IC noise and is better suited for extremely localized defects. In fact, in the presence of small-scale defects, the use of all pairwise geodesic distances can mask the OC condition and may lead to a deterioration in detection performance. Furthermore, detection performance degrades under higher IC noise, which significantly masks defects.

The proposed procedures require high-density point clouds. In our simulations, we used point clouds of tens of thousands of points, corresponding to very small printed parts. However, in practice, collected point clouds typically range from hundreds of thousands to millions of points. Future developments of the procedure include post-signal diagnostics to localize defects, identifying both their location and spatial spread, once an alarm is triggered. Additional research could be conducted to simultaneously monitor the shape and point-wise attributes, such as surface texture.

# A Appendix

## A.1 On choosing the number of eigenvalues for RFM-RL

The number of eigenvalues $k$ can be selected using a elbow rule heuristic, as in classical principal component analysis. In particular, An et al. (2025) presented a method to determine $k$ by measuring the distance - in terms of Frobenius norm - between the CAD model and the reconstructed mesh. The latter can be obtained by projecting the original CAD model coordinates into the subspace spanned by the first $k$ eigenvectors. Specifically, An et al. (2025) applied Gram-Schmidt to ortogonalize eigenvectors of the discrete LB operator. Indeed, the eigenvectors are not orthogonal according to the usual definition (i.e., $\langle \boldsymbol{u}_i, \boldsymbol{u}_j \rangle = \boldsymbol{u}_i^\top \boldsymbol{u}_j \neq 0$, $i \neq j$). However, they are orthogonal with respect to the $\mathbf{M}$-inner product $\langle \boldsymbol{u}_i, \boldsymbol{u}_j \rangle_{\mathbf{M}} = \boldsymbol{u}_i^\top \mathbf{M} \boldsymbol{u}_j = 0$, $i \neq j$, where $\mathbf{M}$ is the mass matrix in the generalized eigenvalue problem in Eq. (2) (see, e.g., Rustamov, 2007; Reuter et al., 2009). Hence, we can avoid applying Gram-Schmidt to achieve orthogonality in the standard Euclidean space and compute the projection with respect to the manifold. In addition, we suggest applying the following procedure to available or generated IC samples, rather than to the CAD model. This ensures that $k$ is selected accounting for the IC variability.

Let $\mathbf{S}_0 \in \mathbb{R}^{n\times 3}$ denotes the point cloud of an IC sample and $\mathbf{U}_k \in \mathbb{R}^{n\times k}$ be the matrix whose columns are the first $k$ eigenvectors of the generalized eigenvalue problem in Eq. (2). Our aim is to minimize

$$\min_{\mathbf{S}\in \text{span}(\mathbf{U}_k)} \|\mathbf{S}_0 - \mathbf{S}\|_{\mathbf{M}}^2$$

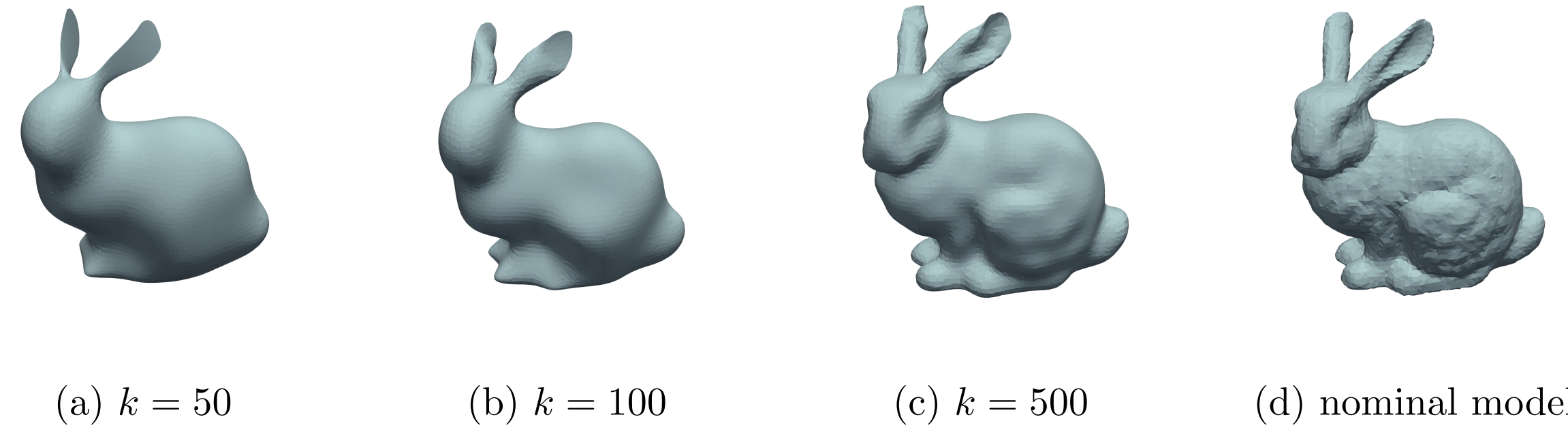

(a) $k = 50$ (b) $k = 100$ (c) $k = 500$ (d) nominal model

Figure 10: Reconstruction of the nominal model using the first $k$ eigenvectors of RL.

where $\mathbf{S}$ can be expressed as $\mathbf{S} = \mathbf{U}_k \mathbf{S}_k$. Hence, we want to minimize

$$\operatorname{tr}\left[(\mathbf{S}_0 - \mathbf{U}_k\mathbf{S}_k)^\top \mathbf{M}\,(\mathbf{S}_0 - \mathbf{U}_k\mathbf{S}_k)\right] = \operatorname{tr}\left[\mathbf{S}_0^\top \mathbf{M}\mathbf{S}_0 - 2\mathbf{S}_0^\top \mathbf{M}\mathbf{U}_k\mathbf{S}_k + \mathbf{S}_k^\top \mathbf{U}_k^\top \mathbf{M}\mathbf{U}_k\mathbf{S}_k\right]$$

Then, by taking the derivative with respect to $\mathbf{S}_k$ and equating it to zero, we obtain $\mathbf{S}_k = \left(\mathbf{U}_k^\top \mathbf{M}\mathbf{U}_k\right)^{-1} \mathbf{U}_k^\top \mathbf{M}\mathbf{S}_0$. Since $\mathbf{S} = \mathbf{U}_k\mathbf{S}_k$, this leads to

$$\mathbf{S} = \mathbf{U}_k \left(\mathbf{U}_k^\top \mathbf{M}\mathbf{U}_k\right)^{-1} \mathbf{U}_k^\top \mathbf{M}\mathbf{S}_0$$

Therefore, a measure of discrepancy between an IC sample and the point cloud reconstructed using the first $k$ eigenvectors is given by $D = \|\mathbf{S}_0 - \mathbf{S}\|_F$. By plotting $k$ against $D$, the elbow rule heuristic can be used to select an appropriate value for $k$.

## A.2 Control limit estimation

We describe the permutation-based approach for estimating the control limit for both RFM-RL and RFM-HM. Our algorithm quantifies estimation uncertainty in the IC parameters, control limit, and attained ARL. In each permutation $\ell$, the reference sample is permuted and divided into three sets of sizes $m_1$, $m_2$ and $m_3$: *training*, *tuning* and *test*. The *training* set is used to estimate $\boldsymbol{\mu}^{(\ell)} \in \mathbb{R}^k$ and $\boldsymbol{\Sigma}^{(\ell)} \in \mathbb{R}^{k\times k}$. The observations of the *tuning* set are standardized using these estimates, control statistics are computed, and the empirical quantile $q_{1-\alpha}^{(\ell)}$ is determined (i.e., the control limit). The *test* set is then standardized, control statistics are computed, and the number $a^{(\ell)}$ of statistics exceeding $q_{1-\alpha}^{(\ell)}$ is recorded, yielding $\text{ARL}^{(\ell)} = m_3/a^{(\ell)}$. Based on our empirical findings, we recommend at least $L = 1000$ permutations. This permutation-based approach yields a distribution for the IC parameters, control limit, and attained ARL across the $L$ permutations. For OC performance evaluation, only the first two sets are permuted and $\text{ARL}^{(\ell)}$ is computed from the number of control statistics (from the OC set) that exceed $q_{1-\alpha}^{(\ell)}$.

## Acknowledgments

The authors thank Professor Bianca Maria Colosimo and Professor Marco Grasso from Politecnico di Milano for providing the data used in the case study. This research was supported in part through research cyberinfrastructure resources and services provided by the Partnership for an Advanced Computing Environment (PACE) at the Georgia Institute of Technology, Atlanta, Georgia, USA.

## Disclosure of Interest

The authors declare no conflicts of interest.

## Data Availability Statement

The code and data required to reproduce all the results presented in this paper are publicly available at https://github.com/franci2312/RFM.

# References

Amenta, N. and Bern, M. (1999). Surface reconstruction by voronoi filtering. *Discrete and Computational Geometry*, 22:481–504.

An, Y., Zhao, X., and Castillo, E. D. (2025). Practical implementation of an end-to-end methodology for spc of 3-D part geometry: A case study. *Journal of Quality Technology*, 57(4):366–381.

Belkin, M., Sun, J., and Wang, Y. (2009). Constructing laplace operator from point clouds in Rd. In *Proceedings of the Twentieth Annual ACM-SIAM Symposium on Discrete Algorithms*, SODA '09, page 1031–1040, USA. Society for Industrial and Applied Mathematics.

Bernardini, F., Mittleman, J., Rushmeier, H., Silva, C., and Taubin, G. (1999). The ball-pivoting algorithm for surface reconstruction. *IEEE Transactions on Visualization and Computer Graphics*, 5(4):349–359.

Bernstein, M., de Silva, V., Langford, J. C., and Tenenbaum, J. B. (2000). Graph approximations to geodesics on embedded manifolds. Technical report, Department of Psychology, Stanford University.

Besl, P. and McKay, N. D. (1992). A method for registration of 3-D shapes. *IEEE Transactions on Pattern Analysis and Machine Intelligence*, 14(2):239–256.

Bobenko, A. I. and Springborn, B. A. (2007). A discrete laplace—beltrami operator for simplicial surfaces. *Discrete Comput. Geom.*, 38(4):740–756.

Bronstein, M. M., Bruna, J., LeCun, Y., Szlam, A., and Vandergheynst, P. (2017). Geometric deep learning: Going beyond euclidean data. *IEEE Signal Processing Magazine*, 34(4):18–42.

Chung, F. R. K. (1997). *Spectral Graph Theory*. CBMS Regional Conference Series in Mathematics, No. 92. American Mathematical Society.

Clarenz, U., Rumpf, M., and Telea, A. (2004). Finite Elements on Point Based Surfaces. In Gross, M., Pfister, H., Alexa, M., and Rusinkiewicz, S., editors, *SPBG'04 Symposium on Point - Based Graphics 2004*. The Eurographics Association.

Colosimo, B. M., Garghetti, F., Grasso, M., and Pagani, L. (2024). On-line inspection of lattice structures and metamaterials via in-situ imaging in additive manufacturing. *Additive Manufacturing*, 95:104538.

Colosimo, B. M., Grasso, M., Garghetti, F., and Rossi, B. (2022). Complex geometries in additive manufacturing: A new solution for lattice structure modeling and monitoring. *Journal of Quality Technology*, 54(4):392–414.

Crane, K., Livesu, M., Puppo, E., and Qin, Y. (2020). A Survey of Algorithms for Geodesic Paths and Distances. *arXiv e-prints*, page arXiv:2007.10430.

Crane, K., Weischedel, C., and Wardetzky, M. (2017). The heat method for distance computation. *Commun. ACM*, 60(11):90–99.

Del Castillo, E., Colosimo, B. M., and Tajbakhsh, S. D. (2015). Geodesic gaussian processes for the parametric reconstruction of a free-form surface. *Technometrics*, 57(1):87–99.

Del Castillo, E. and Zhao, X. (2020). Industrial statistics and manifold data. *Quality Engineering*, 32(2):155–167.

Dijkstra, E. W. (1959). A note on two problems in connexion with graphs. *Numerische Mathematik*, 1:269–271.

Donoho, D. L. and Johnstone, I. M. (1994). Ideal spatial adaptation by wavelet shrinkage. *Biometrika*, 81(3):425–455.

Edelsbrunner, H., Kirkpatrick, D., and Seidel, R. (1983). On the shape of a set of points in the plane. *IEEE Transactions on Information Theory*, 29(4):551–559.

Ertas, A., Farley-Talamantes, E., Cuvalci, O., and Gecgel, O. (2025). 3D-printing of artificial aortic heart valve using uv-cured silicone: Design and performance analysis. *Bioengineering*, 12(1).

Fan, J. (1996). Test of significance based on wavelet thresholding and neyman's truncation. *Journal of the American Statistical Association*, 91(434):674–688.

Haghiashtiani, G., Qiu, K., Sanchez, J. D. Z., Fuenning, Z. J., Nair, P., Ahlberg, S. E., Iaizzo, P. A., and McAlpine, M. C. (2020). 3D printed patient-specific aortic root models with internal sensors for minimally invasive applications. *Science Advances*, 6(35):eabb4641.

Hoppe, H., DeRose, T., Duchamp, T., McDonald, J., and Stuetzle, W. (1993). Mesh optimization. In *Proceedings of the 20th Annual Conference on Computer Graphics and*

*Interactive Techniques*, SIGGRAPH '93, page 19–26, New York, NY, USA. Association for Computing Machinery.

Hron, K., Menafoglio, A., Templ, M., Hrůzová, K., and Filzmoser, P. (2016). Simplicial principal component analysis for density functions in bayes spaces. *Computational Statistics Data Analysis*, 94:330–350.

Huang, D., Du, S., Li, G., Zhao, C., and Deng, Y. (2018). Detection and monitoring of defects on three-dimensional curved surfaces based on high-density point cloud data. *Precision Engineering*, 53:79–95.

Jeong, M. K., Lu, J.-C., and Wang, N. (2006). Wavelet-based SPC procedure for complicated functional data. *International Journal of Production Research*, 44(4):729–744.

Jeong, M. K., Lu, J.-C., Zhou, W., and Ghosh, S. K. (2007). Data-reduction method for spatial data using a structured wavelet model. *International journal of production research*, 45(10):2295–2311.

Kalender, W. A. (2011). *Computed tomography: fundamentals, system technology, image quality, applications*. John Wiley & Sons.

Kazhdan, M., Bolitho, M., and Hoppe, H. (2006). Poisson surface reconstruction. In *Proceedings of the fourth Eurographics symposium on Geometry processing*, volume 7.

Kazhdan, M. and Hoppe, H. (2013). Screened poisson surface reconstruction. *ACM Transactions on Graphics (TOG)*, 32(3):29:1–29:13.

Kimmel, R. and Sethian, J. A. (1998). Computing geodesic paths on manifolds. *Proceedings of the national academy of Sciences*, 95(15):8431–8435.

Lanthier, M., Maheshwari, A., and Sack, J.-R. (1997). Approximating weighted shortest paths on polyhedral surfaces. In *Proceedings of the Thirteenth Annual Symposium on Computational Geometry*, SCG '97, page 485–486, New York, NY, USA. Association for Computing Machinery.

Lehoucq, R. B., Sorensen, D. C., and Yang, C. (1998). *ARPACK Users' Guide*. Society for Industrial and Applied Mathematics.

Levy, B. (2006). Laplace-beltrami eigenfunctions towards an algorithm that understands geometry. volume 2006, pages 13 – 13.

Li, X., Xu, G., and Zhang, Y. J. (2015). Localized discrete laplace–beltrami operator over triangular mesh. *Computer Aided Geometric Design*, 39:67–82.

Liu, Y., Prabhakaran, B., and Guo, X. (2012). Point-based manifold harmonics. *IEEE Transactions on Visualization and Computer Graphics*, 18(10):1693–1703.

Mahalanobis, P. C. (1936). On the generalised distance in statistics. *Proceedings of the National Institute of Sciences of India*, 2:49–55.

Mémoli, F. and Sapiro, G. (2005). Distance functions and geodesics on submanifolds of $\mathbb{R}^d$ and point clouds. *SIAM Journal on Applied Mathematics*, 65(4):1227–1260.

Menafoglio, A., Grasso, M., Secchi, P., and Colosimo, B. M. (2018). Profile monitoring

of probability density functions via simplicial functional pca with application to image data. *Technometrics*, 60(4):497–510.

Mitchell, J. S. B., Mount, D. M., and Papadimitriou, C. H. (1987). The discrete geodesic problem. *SIAM J. Comput.*, 16:647–668.

Muntoni, A., Cignoni, P., Luaces, A., and Zhang, F. (2025). Pymeshlab: Python bindings for meshlab.

Mwema, F. M. and Akinlabi, E. T. (2020). Basics of fused deposition modelling (fdm). *Fused Deposition Modeling Strategies for Quality Enhancement*, page 1—15.

Niethammer, M., Reuter, M., Wolter, F.-E., Bouix, S., Peinecke, N., Koo, M.-S., and Shenton, M. E. (2007). Global medical shape analysis using the laplace-beltrami spectrum. In Ayache, N., Ourselin, S., and Maeder, A., editors, *Medical Image Computing and Computer-Assisted Intervention – MICCAI 2007*, pages 850–857, Berlin, Heidelberg. Springer Berlin Heidelberg.

Patane, G. (2017). An introduction to laplacian spectral distances and kernels: Theory, computation, and applications. *Synthesis Lectures on Visual Computing*, 9:1–139.

Reuter, M., Biasotti, S., Giorgi, D., Patanè, G., and Spagnuolo, M. (2009). Discrete laplace–beltrami operators for shape analysis and segmentation. *Computers Graphics*, 33(3):381–390.

Reuter, M., Wolter, F.-E., and Peinecke, N. (2006). Laplace–beltrami spectra as 'shape-

dna' of surfaces and solids. *Computer-Aided Design*, 38(4):342–366. Symposium on Solid and Physical Modeling 2005.

Roweis, S. T. and Saul, L. K. (2000). Nonlinear dimensionality reduction by locally linear embedding. *Science*, 290(5500):2323–2326.

Rustamov, R. M. (2007). Laplace-Beltrami Eigenfunctions for Deformation Invariant Shape Representation. In Belyaev, A. and Garland, M., editors, *Geometry Processing*. The Eurographics Association.

Scimone, R., Taormina, T., Colosimo, B. M., Grasso, M., Menafoglio, A., and Secchi, P. (2022). Statistical modeling and monitoring of geometrical deviations in complex shapes with application to additive manufacturing. *Technometrics*, 64(4):437–456.

Senin, N., Catalucci, S., Moretti, M., and Leach, R. (2021). Statistical point cloud model to investigate measurement uncertainty in coordinate metrology. *Precision Engineering*, 70:44–62.

Sharp, N. and Crane, K. (2020). A Laplacian for Nonmanifold Triangle Meshes. *Computer Graphics Forum (SGP)*, 39(5).

Stankus, S. E. and Castillo-Villar, K. K. (2019). An improved multivariate generalised likelihood ratio control chart for the monitoring of point clouds from 3D laser scanners. *International Journal of Production Research*, 57(8):2344–2355.

Surazhsky, V., Surazhsky, T., Kirsanov, D., Gortler, S. J., and Hoppe, H. (2005). Fast

exact and approximate geodesics on meshes. *ACM transactions on graphics (TOG)*, 24(3):553–560.

Tenenbaum, J. B., de Silva, V., and Langford, J. C. (2000). A global geometric framework for nonlinear dimensionality reduction. *Science*, 290(5500):2319–2323.

van der Maaten, L. and Hinton, G. (2008). Visualizing data using t-sne. *Journal of Machine Learning Research*, 9(86):2579–2605.

Wang, A., Wang, K., and Tsung, F. (2014). Statistical surface monitoring by spatial-structure modeling. *Journal of Quality Technology*, 46(4):359–376.

Wang, Y., Mei, Y., and Paynabar, K. (2018). Thresholded multivariate principal component analysis for phase I multichannel profile monitoring. *Technometrics*, 60(3):360–372.

Wardetzky, M., Mathur, S., Kaelberer, F., and Grinspun, E. (2007). Discrete Laplace operators: No free lunch. In Belyaev, A. and Garland, M., editors, *Geometry Processing*. The Eurographics Association.

Wells, L. J., Megahed, F. M., Niziolek, C. B., Camelio, J. A., and Woodall, W. H. (2013). Statistical process monitoring approach for high-density point clouds. *Journal of Intelligent Manufacturing*, 24:1267–1279.

Zang, Y. and Qiu, P. (2018a). Phase I monitoring of spatial surface data from 3D printing. *Technometrics*, 60(2):169–180.

Zang, Y. and Qiu, P. (2018b). Phase II monitoring of free-form surfaces: An application to 3D printing. *Journal of Quality Technology*, 50(4):379–390.

Zhao, X. and Del Castillo, E. (2021). An intrinsic geometrical approach for statistical process control of surface and manifold data. *Technometrics*, 63(3):295–312.

Zhao, X. and Del Castillo, E. (2022). A registration-free approach for statistical process control of 3D scanned objects via FEM. *Precision Engineering*, 74:247–263.

Zou, C., Ning, X., and Tsung, F. (2010). LASSO-based multivariate linear profile monitoring. *Annals of Operations Research*, 192:3–19.